\documentclass[conference]{IEEEtran}% 
\IEEEoverridecommandlockouts
\usepackage{cite}
\usepackage{footmisc}
\usepackage{amsmath,amssymb,amsfonts}
\usepackage{graphicx}
\usepackage{textcomp}
\usepackage{xcolor}
\usepackage{mathtools} % 
\usepackage{todonotes} %
\usepackage[caption=false]{subfig} %
\usepackage[mathscr]{euscript} % euler, eucal
\usepackage{url} %
\usepackage{hyperref}
\usepackage{bm}
\usepackage{algpseudocode} 
\usepackage{algorithm}
\usepackage{xifthen}
\usepackage[export]{adjustbox}
\usepackage{textcomp}
\usepackage{array}   
\usepackage{svg}
\usepackage{enumitem}
\usepackage{xcolor}
\usepackage{wrapfig} % images wrapped by text
\usepackage{colortbl}  % Sometimes helpful for older LaTeX installations
\newcommand{\mycomment}[1]{}
\NewDocumentCommand{\vect}{ O{} O{} m }{\mathbf{#3}\ifthenelse{\isempty{#1}}{}{^{(#1)}}\ifthenelse{\isempty{#2}}{}{_{#2}}}
\NewDocumentCommand{\mat}{ O{} O{} m }{\mathbf{#3}\ifthenelse{\isempty{#1}}{}{^{(#1)}}\ifthenelse{\isempty{#2}}{}{_{#2}}}
\NewDocumentCommand{\ten}{ O{} O{} m }{\pmb{\mathscr{#3}}\ifthenelse{\isempty{#1}}{}{^{(#1)}}\ifthenelse{\isempty{#2}}{}{_{#2}}}

\def\BibTeX{{\rm B\kern-.05em{\sc i\kern-.025em b}\kern-.08em
    T\kern-.1667em\lower.7ex\hbox{E}\kern-.125emX}}
\definecolor{mygreen}{rgb}{0,0.6,0}
\definecolor{mymauve}{rgb}{0.58,0,0.82}
\definecolor{mygray}{rgb}{0.5,0.5,0.5}
\definecolor{blue}{rgb}{0.0,0.0,1.0}
\definecolor{red}{rgb}{1.0,0.0,0.0}
\usepackage{listings}
\lstdefinelanguage{cypher2}{
    sensitive=true,
    morekeywords=[1]{MATCH, RETURN, WHERE, CONTAINS},
    morekeywords=[2]{PERSON, FRIEND,Document, Keyword, Affiliation, Country},
    morestring=[b]",
    morecomment=[l]{//},
    morecomment=[s]{/*}{*/},
    morecomment=[s]{--}{\ },
}
\lstdefinestyle{cypherstyle2}{
    language=cypher2,
    basicstyle=\footnotesize\ttfamily,
    keywordstyle=\color{blue}\bfseries, 
    keywordstyle=[2]\color{red}\bfseries,   
    commentstyle=\color{mygreen},
    stringstyle=\color{mymauve},
    numberstyle=\tiny\color{mygray},
    breaklines=true,
    showstringspaces=false,
    captionpos=b
}

\usepackage{listings}
\begin{document}
\title{Domain-Specific Retrieval-Augmented Generation Using Vector Stores, Knowledge Graphs, and Tensor Factorization
}

\author{\IEEEauthorblockN{
Ryan C. Barron\IEEEauthorrefmark{2}\IEEEauthorrefmark{3}\IEEEauthorrefmark{6},
Vesselin Grantcharov\IEEEauthorrefmark{4}\IEEEauthorrefmark{6}, 
Selma Wanna\IEEEauthorrefmark{1}\IEEEauthorrefmark{5},
Maksim E. Eren\IEEEauthorrefmark{1}\IEEEauthorrefmark{3},\\
Manish Bhattarai\IEEEauthorrefmark{2},
Nicholas Solovyev\IEEEauthorrefmark{2},
George Tompkins\IEEEauthorrefmark{7},\\
Charles Nicholas\IEEEauthorrefmark{3}\IEEEauthorrefmark{1},
Kim {\O}.  Rasmussen\IEEEauthorrefmark{2},
Cynthia Matuszek\IEEEauthorrefmark{3}\IEEEauthorrefmark{1},
and Boian S. Alexandrov\IEEEauthorrefmark{2}
}
\IEEEauthorblockA{
\IEEEauthorrefmark{7}Analytics, Intelligence and Technology Division, Los Alamos National Laboratory, Los Alamos, New Mexico, USA. \\
\IEEEauthorrefmark{4}University of New Mexico.
\IEEEauthorrefmark{5}University of Texas at Austin.
\IEEEauthorrefmark{3}University of Maryland Baltimore County.\\
\IEEEauthorrefmark{1}Advanced Research in Cyber Systems, Los Alamos National Laboratory, New Mexico, USA. \\
\IEEEauthorrefmark{2}Theoretical Division, Los Alamos National Laboratory, New Mexico, USA. 
}

\thanks{\IEEEauthorrefmark{6}The first two authors contributed equally to this work.}
\thanks{U.S. Government work not protected by U.S. \%copyright.}
}

\maketitle
\begin{abstract}
Large Language Models (LLMs) are pre-trained on large-scale corpora and excel in numerous general natural language processing (NLP) tasks, such as question answering (QA). Despite their advanced language capabilities, when it comes to domain-specific and knowledge-intensive tasks, LLMs suffer from hallucinations, knowledge cut-offs, and lack of knowledge attributions. Additionally, fine tuning LLMs' intrinsic knowledge to highly specific domains is an expensive and time consuming process. The retrieval-augmented generation (RAG) process has recently emerged as a method capable of optimization of LLM responses, by referencing them to a predetermined ontology. It was shown that using a Knowledge Graph (KG) ontology for RAG improves the QA accuracy, by taking into account relevant sub-graphs that preserve the information in a structured manner. In this paper, we introduce SMART-SLIC, a highly domain-specific LLM framework, that integrates RAG with KG and a vector store (VS) that store factual domain specific information. Importantly, to avoid hallucinations in the KG, we build these highly domain-specific KGs and VSs without the use of LLMs, but via NLP, data mining, and nonnegative tensor factorization with automatic model selection. Pairing our RAG with a domain-specific: (i) KG (containing structured information), and (ii) VS (containing unstructured information) enables the development of domain-specific chat-bots that attribute the source of information, mitigate hallucinations, lessen the need for fine-tuning, and excel in highly domain-specific question answering tasks. We pair SMART-SLIC with chain-of-thought prompting agents. The framework is designed to be generalizable to adapt to any specific or specialized domain. In this paper, we demonstrate the question answering capabilities of our framework on a corpus of scientific publications on malware analysis and anomaly detection. 
\end{abstract}

\begin{IEEEkeywords}
Artificial Intelligence, Retrieval Augmented Generation, Knowledge Graph, Natural Language Processing, Non-Negative Tensor Factorization, Topic Modeling, Agents
\end{IEEEkeywords}

\section{Introduction}
\label{sec:introduction}

The expanding volumes of data across large databases and information collections necessitate the specialized extraction of pertinent knowledge, often without an in-depth understanding of the underlying database resources. Recent advancements in Large Language Models (LLMs) have facilitated developments that enable users to engage in dialogues with LLM-powered chat-bots to discover information. Despite these models' impressive handling of general queries, their application in domain-specific tasks is hindered by several limitations. These include the production of factually incorrect responses ("hallucinations") \cite{yadkori2024believe}, unawareness of recent developments or events beyond their training data ("knowledge cutoff") \cite{harvel2024can}, failure to accurately attribute sources of information ("implicit knowledge") \cite{zhang2024comprehensive}, and a lack of specific technical knowledge required for specialized fields \cite{freire2024chatbots}.

Fine-tuning is a common strategy employed to tailor these general models to specific domains. However, this approach is resource-intensive, demanding significant amounts of data, extensive computational power, and considerable time, which makes it impractical for many domain-specific applications. These limitations pose significant challenges in interpreting and validating the knowledge generated by LLMs, as well as in referencing their sources. Consequently, this reduces the trustworthiness of LLMs and limits their effectiveness in highly specialized scientific contexts where accuracy and reliability are paramount. The ongoing challenges underscore the need for more sophisticated solutions that can bridge the gap between general-purpose LLMs and the nuanced requirements of domain-specific applications.

% Xu2024} xu2024retrieval
Retrieval-Augmented Generation (RAG) with Knowledge Graphs (KGs) and vector stores (VS) significantly enhances the context of LLMs, mitigating the need to fine-tune these models to specific domains \cite{bertsch2024context, lewis2020retrieval}. KGs provide a structured way to store factual information, making it easier to access and use, while VSs allow storing unstructured documents and preserving the semantics of the text. This integration allows LLMs to tap into both domain-specific and updated information, effectively addressing the traditional limitations of generative models.

Despite these improvements, challenges remain in the practical implementation of domain-specific RAG systems. Extracting accurate and representative domain-specific ontologies to build KGs and VSs is a complex task. Additionally, curating datasets with specific text data for constructing both KGs and VSs is equally demanding. These steps are critical for ensuring that the augmented LLMs can reliably produce high-quality, relevant responses across different domains.

In this paper, 
% building upon previous work\footnote{\url{https://smart-tensors.lanl.gov/}}, 
we introduce a framework designed for constructing domain-specific corpora of scientific articles through advanced techniques, including: text mining, information retrieval, dimension reduction, nonnegative tensor factorization, citation graphs, and human-in-the-loop strategies. We introduce a novel framework, which we call \textbf{SMART-SLIC},
%(for \textbf{SLIC} pls. see Ref.\cite{eren2023slic})
for developing KG's ontologies, utilizing both metadata and full texts from open-source scientific publications, as well the latent structures of these corpora, extracted through nonnegative tensor factorization, enhanced with automatic model determination. \textbf{SMART-SLIC} facilitates topic modeling \cite{Vangara2020}, 
and determination of the optimal number of topics \cite{nebgen2021neural,vangara2021finding} 
for effective document classification. 
Our new framework underpins the creation of a precisely tailored corpus of domain-specific scientific articles, which is crucial for our AG approach and supports the development of a chat-bot adept at answering domain-specific technical inquiries. Further, the framework is versatile, allowing for its application to any domain of documents. In this paper, we illustrate the effectiveness of our framework, \textbf{SMART-SLIC}, with a case study where we construct a domain-specific corpus, KG, and VS, focused on malware analysis and anomaly detection, and apply our enhanced question-answering framework for scientific queries related to this corpus. Our contributions are summarized as follows:

\begin{itemize}
    \item We detail the development of a framework for building domain-specific scientific corpora using a blend of text mining, information retrieval, artificial intelligence (AI), and human-in-the-loop techniques.
    \item We describe the creation of a domain-specific KG $\&$ VS ontology that leverages both observable metadata, and full texts of the corpus of domain-specific open-source scientific articles, as well as its latent structure extracted by non-negative tensor/matrix factorization with automatic model selection.
    \item We demonstrate the enhanced capabilities of \textbf{SMART-SLIC}'s, RAG-enhanced LLM system, which utilizes chain-of-thought prompting with LLM agents to proficiently address scientific questions.
\end{itemize}

\section{Related Works}
\label{sec:related_works}
Recent methods for building RAG-assisted \cite{lewis2020retrieval} chatbot applications rely on unstructured text stored in vector databases for question answering (QA) tasks \cite{Liu_LlamaIndex_2022}. Although the integration of knowledge graphs (KGs) in AI systems is not novel \cite{kgqa_old}, increasingly, researchers are leveraging them to improve LLM reasoning while simultaneously addressing the reliability issues discussed in Section \ref{sec:introduction} \cite{ VLNSurvey_Pan1, VLNSurvey_Pan2, VLNSurvey_Li}. Despite the benefits, integrating domain-specific knowledge into chatbots requires substantial effort. Here, we review the prior work for common chatbot designs, the integration of domain-knowledge in RAG pipelines, and the steps required for constructing KGs.

\subsection{KGs in RAG Pipelines}  
Building a sophisticated chatbot requires the knowledge of a wide range of research fields; hence, rarely do prior works present a fully engineered system like ours. Instead, most efforts focus on improving specific aspects of RAG pipelines, e.g., retriever design \cite{chen-etal-2017-reading, karpukhin-etal-2020-dense}, query intent recognition \cite{slimproxy}, and KG reasoning \cite{ jiang-etal-2023-reasoninglm, luo2024reasoning, jin2024graphchainofthoughtaugmentinglarge, li2024kgmistral}. Our approach resembles past methods which leverage chain-of-thought \cite{wei2023chainofthoughtpromptingelicitsreasoning} prompting on KGs \cite{jin2024graphchainofthoughtaugmentinglarge, luo2024reasoning}; in conjunction with LLM-agents to enhance reasoning capabilities \cite{yao2023react, lala2023paperqa, sanmartin2024kgragbridginggapknowledge}. In addition to incorporating these state-of-the-art techniques, we improve our RAG pipeline by modifying our retrieval method to use K-Nearest Neighbors with the Levenshtein metric instead of cosine distance as an entry point for context search. We also construct a ``highly-specific" knowledge base for targeted QA tasks.

% \subsection{Domain-Specific KGs in RAG Pipelines} 
Although expensive and time-consuming, a handful of prior works incorporate domain-knowledge into their RAG pipelines \cite{soman2023biomedical, edwards2024hybrid, kragen, Xu2024}; however, the majority either use existing KGs built broadly on medical literature \cite{soman2023biomedical, kragen}; or do not disclose any details regarding their dataset construction \cite{Xu2024}. We emphasize that our method is ``highly-specific" because it was driven by subject matter expertise which informed our dataset curation and cleaning techniques \cite{eren2023slic, TELF}. 

\subsection{KG Development}
At a minimum, the development of knowledge graphs requires building a corpus, defining an ontology, and extracting the relevant entity-relation triplets from unstructured text. 

\textbf{Corpus Building.} Here we define the term ``highly-specific" and explain our dataset collection method. A key feature of our dataset collection is the use of unsupervised methods \cite{TELF} to decompose corpora into document clusters to finer specificity than the author-provided tags available on open access websites. This differs significantly from prior approaches \cite{ABUSALIH2021103076, edwards2024hybrid, yu2022biosalgorithmicallygeneratedbiomedical}. We leverage latent-topic information from our NMFk method to filter and select the best data for our knowledge base, and prune documents based on citation information and embedding distances. Our text cleaning pipeline is informed by subject matter experts (SME)  \cite{TELF, 10460022}, thus going beyond standard methods by incorporating expert-derived rules for document cleaning, e.g, acronym and entity standardization. 

\textbf{KG Construction.} Our ontology is shaped by traditional methods, i.e., relying on SME design and capturing task-specific features. However, we innovate by incorporating latent information from our decomposition process \cite{TELF} into our KG as entities. For entity and relation extraction, we move beyond conventional learning-based techniques \cite{ji2022}; and instead, leverage recent advancements which use LLM-agents \cite{Liu_LlamaIndex_2022, ye2023isolationmultiagentsynergyimproving} as opposed to other LLM prompting methods \cite{wadhwa-etal-2023-revisiting, marinov2024relation, zavarella2024a, edge2024localglobalgraphrag}. This approach yields non-sparse KGs, meaning, the average out-degree of entities \cite{lv-etal-2020-dynamic, HoGRN} is high. To our knowledge, no prior work integrates all of these methods into their knowledge graph construction process.

 \section{Methods}
\label{sec:methods}
This section outlines our framework, covering corpus extraction, KG ontology, VS construction, and the RAG process.

\subsection{Domain-Specific Dataset}

\begin{wrapfigure}{l}{0.5\columnwidth}
    \centering
    \includegraphics[width=0.5\columnwidth]{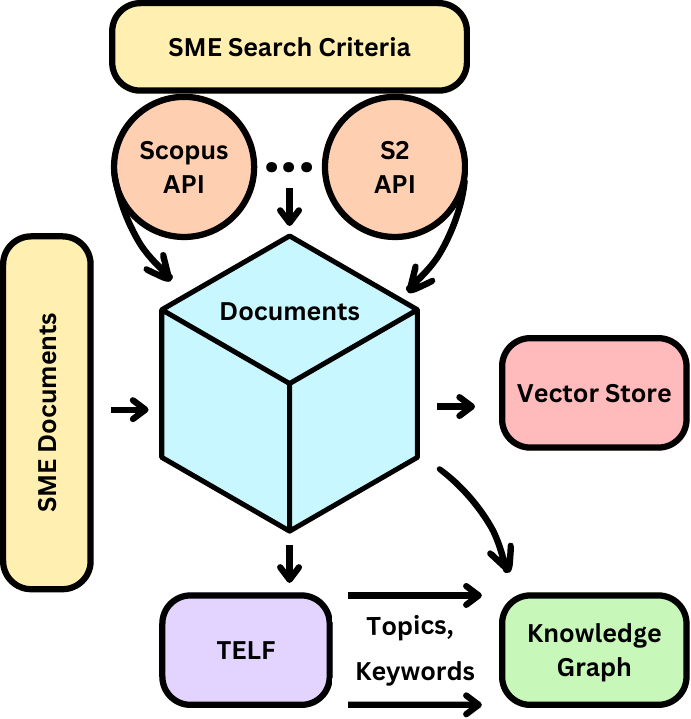}
    \vspace{-2em}
    \caption{ User query routing overview.}
    \label{fig:data_construct}
\end{wrapfigure}

Overview of of our system is summarized in Figure \ref{fig:data_construct}. To collect the dataset, we began with a set of core documents selected by subject matter experts (SMEs). 
Here, these core documents represent the specific domain in which we want to built our corpus on. These core documents were used to build a citation and reference network, which allowed for the expansion of the dataset through the authorized APIs: SCOPUS \cite{scopus_website}, Semantic Scholar (S2) \cite{semantic_scholar_website}, and  Office of Scientific and Technical Information (OSTI) \cite{osti_website}. 
We also extract common bigrams from the core documents to query these APIs to search for relevant documents. As we expand on the the corpus starting from the core documents, it is possible to add documents that do not directly relate to the information in the core documents. To maintain the central quality and thematic coherence of the core dataset, we employed several pruning strategies to remove these irrelevant documents to preserve the speciality specific to the targeted domain. These strategies focused on removing documents that diverge from the central theme of the core. Pruning was performed through two methods from \cite{10460022}:
\begin{itemize}
\footnotesize
    \item \textbf{Human-in-the-Loop Pruning}:
SMEs manually review and select a handful documents that align with the core theme. Here, we reduce the document's TF-IDF matrix to two dimensions with UMAP and let the SME look at the documents that are at the centroids of the given clusters. SME can then select which documents to remove.
    \item \textbf{Automatic Pruning of Document Embeddings}:
    Based on the SME selections from the previous step, we next remove the document that are certain distance away from the selected and the core documents. Documents were transformed into embeddings with SCI-NCL \cite{Ostendorff2022scincl}, a BERT based model fine-tuned on scientific literature, to measure semantic similarity with core and SME selected documents. Those outside a set similarity threshold were removed, ensuring only the documents relevant to the core documents and SME selections remained.
\end{itemize} 
Although a human is in the loop, the system remains scalable by clustering documents. One review per cluster allows the operator to decide on all documents in the group, making it efficient even with large datasets without limit on cluster size.

Additionally, we applied pre-processing techniques
using a publicly available Python library, \textbf{T}ensor \textbf{E}xtraction of \textbf{L}atent \textbf{F}eatures (\textbf{T-ELF})\footnote{T-ELF is available at \url{https://github.com/lanl/T-ELF}} \cite{TELF}. The cleaning procedures involved the following pre-processing steps:
\begin{itemize}
\footnotesize
    \item Exclude non-English, copyrights, and non-essential elements: stop phrases, formulas, and email addresses.
    \item Remove formatting artifacts like next-line markers, parentheses, brackets, accents, and special characters.
    \item Filter out non-ASCII characters and boundaries, HTML tags, stop words, and standalone numbers.
    \item Eliminate extra whitespace and words $\leq 2$  characters.
    \item Standardize punctuation variations, particularly hyphens.
\end{itemize}

These pre-processing cleaning and standardization efforts are essential for preparing the dataset for further analysis, thereby enhancing the quality and consistency of the data.

\subsection{Dimension Reduction}

The extraction of the latent structure from the dataset is accomplished through the following approach. Initially, the data is prepared and the necessary computational framework is established through these steps:

\begin{itemize}
\item Creation of the TF-IDF matrix, \( \mathbf{X} \), of the cleaned corpus
\item \( \mathbf{X} \) is decomposed using nonnegative tensor factorization from \textbf{T-ELF} enhanced with our new binary search strategy \cite{barron2024binarybleedfastdistributed}, 
to classify document clusters.
\end{itemize}

\textbf{T-ELF}  allows us to extract highly specific features from the data. This method identifies latent topics within the corpus, grouping documents into clusters based on shared themes. To avoid over/under-fitting, automatic model determination is used where the final cluster counts are determined by achieving the highest silhouette scores above a predetermined threshold using the Binary Bleed method \cite{barron2024binarybleedfastdistributed}. 
This method employs a binary search strategy across  \( k \)  values, selectively skipping those  \( k \)  values that do not surpass the silhouette threshold. The search criterion for an optimal  \( k \)  is defined as   \( k_{\text{optimal}} = \max \left\{ k \in \{1, 2, \ldots, K\} : S(f(k)) > T \right\} \), where \( S(f(k)) \) denotes the silhouette score of the \( k \)-th configuration and \( T \) the threshold. Importantly, even after identifying an initial ``optimal" \( k \), higher \( k \) values are visited regardless to ensure no better configuration is overlooked.

The factorization of \( \mathbf{X} \) yields two non-negative factor matrices \( \mathbf{W} \in \mathbb{R}^{m \times k}_{+} \) and \( \mathbf{H} \in \mathbb{R}^{k \times n}_{+} \), ensuring \( X_{ij} \approx \sum_{s} W_{is} H_{sj} \). Distribution of words over topics are captured in \( \mathbf{W} \). The matrix \( \mathbf{H} \) shows the topic distribution across documents, and is used to identify the predominant topic for each document in post-processing. 
Full tensor and matrix factorization implementations of various algorithms are available in \textbf{T-ELF} \footnote{Several tensor and matrix factorization algorithms:  \\ \url{https://github.com/lanl/T-ELF/tree/main/TELF/factorization}}.

\subsection{Knowledge Graph Ontology}

Features from \textbf{T-ELF} and document metadata is mapped into series of head, entity, and tail relations, forming directional triplets, then injected into a  Neo4j \cite{neo4j2023} KG.

Our KG incorporates document metadata as well as the latent features. The primary source of information in the KG comes from documents, which are injected into the graph along with related attributes. Each document node contains information such as DOI, title, abstract, and source API document identifiers. Additional node labels include authors, publication year, Scopus category, affiliations, affiliation country, acronyms, publisher, topics, topic keywords, citations, references, and a subset of NER entities produced from spaCy's NER labels \cite{spacyEnCoreWebTrf}. These NER labels cover events, persons, locations, products, organizations, and geopolitical entities.

The KG nodes represent documents and their associated metadata, while the edges capture the relationships between these entities, such as citations, co-authorships, and topic associations, enabling logical query and retrieval capabilities for the RAG.

\subsection{Vector Store Assembly}

To augment the RAG, we introduced a vector database for the original documents using Milvus \cite{2021milvus}. Additionally, a subset of documents' full texts were vectorized and incorporated into the vector store. Full texts, when available, are segmented into smaller paragraphs, each assigned an integer ID to indicate its position within the original document. These paragraphs are then vectorized through the into embeddings using OpenAI's text-embedding-ada-002 \cite{openai_api} model and imported to the vector store to support the RAG process.

The RAG application can query the vector store to find relevant paragraph chunks from these full texts. 
\begin{wrapfigure}{l}{0.5\columnwidth}
    \centering
    \includegraphics[width=0.5\columnwidth]{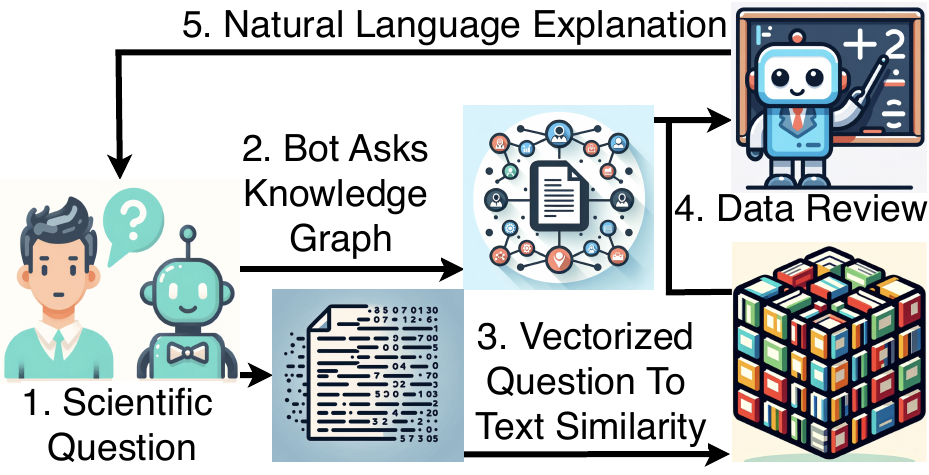}
    \vspace{-2em}
    \caption{  The RAG pipeline. Images generated with DALL·E \cite{dalle_tensor_decomp_arxiv_images}. }
    \label{fig:RAG}
\end{wrapfigure}
If the retrieved text contains the needed information, the LLM can answer the posed question and include a citation of the document, precisely indicating the exact paragraph. If further related information is needed, the application can use document metadata (e.g., DOI, author) to expand its search through the KG. This approach allows us to preserve the semantics of the original documents and provide relevant responses.

\subsection{Retrieval Augmented Generation}

\begin{wrapfigure}{r}{0.53\columnwidth}
    \centering
    \includegraphics[width=0.52\columnwidth]{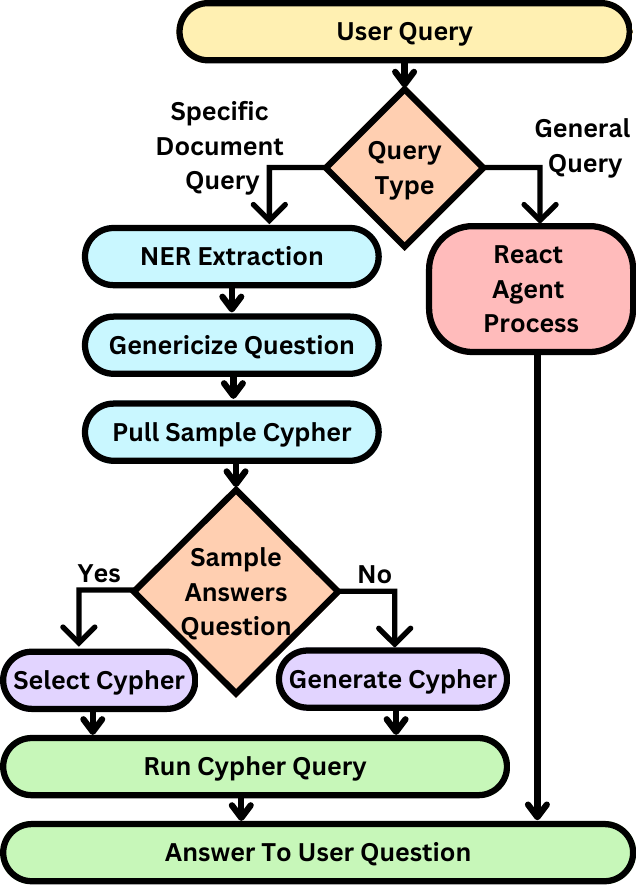}
    \vspace{-2em}
    \caption{ User query routing overview.}
    \label{fig:routing}
\end{wrapfigure}

% RAG is a method in  NLP that mixes retrieval and generation to make answers more accurate and useful in AI systems. Information is first gathered from an outside knowledge source based on a user's question. This information is used to help improve the model's answers, leading to responses that are more relevant and fit the context better. 

% RAG is an NLP technique that combines retrieval and generation methods to enhance response accuracy and relevance in generative AI. RAG operates by first retrieving information from an external knowledge base in response to an input query. This retrieved data is then used to inform and improve the generative model’s outputs, resulting in more contextually relevant responses. 

RAG is an NLP method that mixes retrieval and generation techniques to improve the accuracy and relevance of responses in generative AI. It works by first gathering information from an external knowledge base based on a user’s query. This retrieved information is then used to guide and enhance the outputs of the generative model, leading to more relevant and context-aware responses.
By integrating these tactics, RAG addresses the limitations of purely generative models and provides an adaptable framework suitable for applications demanding detailed and current information. 

Figure \ref{fig:RAG} demonstrates the  data pipeline operated throughout the work for RAG. The process begins with a user query, which the LLM then uses to query the knowledge graph. The LLM transforms the query into a vector embedding. This embedding is compared to existing texts to find the most similar text. The retrieved information is appended to the original query, and the LLM produces a relevant answer using this context. Finally, the LLM constructs a final answer in natural language to explain the answer to the user's question.

%  \begin{figure}[ht]
%     \centering
%     \includegraphics[width=.6 \columnwidth]{figures/user_query.pdf}
%     % \vspace{-2em}
%     \caption{  User query routing overview.}
%     \label{fig:routing}
% \end{figure}

To optimally leverage RAG, accurately understanding the user's question is crucial. 
Our RAG approach includes multiple potential routes depending on a user's question. The question routing pipeline may be a \textbf{General Query}, which calls the \textit{ReAct Agent Process}\cite{yao2023react}, or a \textbf{Specific Document Query}, which calls either a \textit{Retrieved Query} or a \textit{Synthesized Query}. Understanding the question directs the information to the appropriate toolset and subsequent process. The routing process overview, as described below, can be seen in  Figure \ref{fig:routing}.

\textbf{Specific Document Query:} If a user's question requires information from a specific document's text (title + abstract), it is better suited for a traditional RAG application in which the LLM interacts with the VS to find the needed text. In our case, we use a ReAct agent where the VS search is the sole tool, allowing the LLM to make multiple search requests as required. Specifically, a ReAct agent means the LLM has distinct steps for reasoning and acting after determining the input meaning. We use langgraph \cite{langgraph} to define an execution graph with three nodes, as illustrated in Figure \ref{fig:RAG_diagram}: (1) the ReAct agent, (2) the tool executor, and (3) the end.  

\textit{ReAct Agent Process:}  The agent node is the central part of the ReAct graph, where the LLM calls are encapsulated. The ReAct agent is responsible for collecting inputs, making actionable decisions, and explaining the results. The four prompt parts are:
\begin{wrapfigure}{r}{0.5\columnwidth}
    \centering
    \includegraphics[width=0.5\columnwidth]{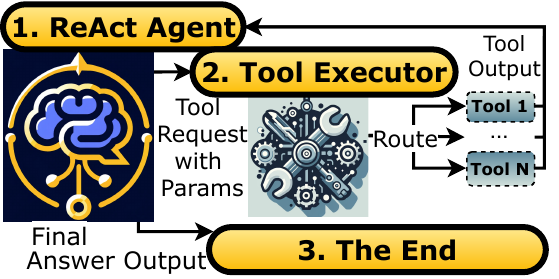}
    \vspace{-2em}
    \caption{ Nodes and tools of the ReAct agent. Images from DALL·E \cite{dalle_tensor_decomp_arxiv_images}.}
    \label{fig:RAG_diagram}
\end{wrapfigure}

\begin{enumerate}[label=\alph*.]
\footnotesize
     \item Instructions 
     \item User query
     \item Tool names, data
     \item Tool Scratchpad
\end{enumerate}

The agent is informed how to answer a user's query from the instructions, including answer formulations and tool usage. The query aids tool selection or answer directly. The tools have specific descriptions and parameters required for their calls, including schemas if interacting with databases. The scratchpad serves as temporary storage for tool calls, responses, and the LLM's reasoning, allowing the agent to iteratively solve complex problems.

The tool executor takes the tool name and input parameters from the agent node, routes to the corresponding function, and returns the output. It handles execution logistics, error handling, logging, and status updates. 

The end node signals that the Reason-Act loop has completed. The final output from the LLM after the retrieval augmented generation is returned to the user. 

% \begin{figure}[ht]
%     \centering
%     \includegraphics[width=\columnwidth]{figures/ReAct_redo.pdf}
%     \vspace{-2em}
%     \caption{ The three execution graph nodes, and the three tools routed from the executor for the ReAct RAG agent. Images generated with DALL·E \cite{dalle_tensor_decomp_arxiv_images}.}
%     \label{fig:RAG_diagram}
% \end{figure}

\textbf{General Query:}  If the user asks a broader question, such as those about trends, the required information is found within the KG. In this route, we start with a preprocessing step in which the LLM performs  NER to decouple specific data from the genericized question. After this, we send the genericized query to a smaller vector store containing pairs of cypher queries and descriptions of the information they return, with embedding vectors generated from the descriptions. From here, there are two possible subroutes. 

\textit{Retrieved Query:} If a retrieved query is able to answer the question, we execute it's cypher before making a final LLM call to return the result. If no existing queries are able to answer, we synthesize a new cypher query.

\textit{Synthesized Query:} If the LLM opts for ``synthesis," it generates a new cypher query using the graph's schema and retrieved examples. For  reliability, the LLM audits this generated query. First, we retrieve the query's execution plan and profile by using the cypher keyword ``PROFILE," which lists the operators used on the knowledge graph. We also provide descriptions of the relevant low-level operators from Neo4j's official documentation. Once we obtain the detailed execution plan, the LLM performs two steps: it translates the plan into plain language and assesses if it addresses the user's question. Valid generations proceed as if retrieved queries.

 \section{Results}
\label{sec:results}
In this section, we discuss identification of optimal clusters for tensor decomposition, vectorization of the dataset, construction of KG, and compare the system using the with GPT-4-instruct \cite{openai_api}  as the operating model of \textbf{SMART-SLIC} to answer research questions. The same model was used to answer without RAG as well. Our findings highlight the accuracy and reliability of the \textbf{SMART-SLIC}'s RAG.

\newcolumntype{P}[1]{>{\centering\arraybackslash}p{#1}}
\begin{table}[ht]
    \centering
    \caption{Labels for Topic Clusters}
    \label{tab:topic_clusters}
    \begin{tabular}{|p{0.02\linewidth}|P{0.56\linewidth}|P{0.1 \linewidth}|P{0.08\linewidth}|}
        \hline
        \#&Label&\# Docs.&Percent\\
        \hline
        \cellcolor{green!10}0 &\cellcolor{green!10} Malware Behavioral Analysis &\cellcolor{green!10} 158 & \cellcolor{green!10}1.80\\
        1 & Cybersecurity Challenges & 305 & 3.47 \\
        \cellcolor{green!10}2 &\cellcolor{green!10} Cybersecurity Research  &\cellcolor{green!10} 114 & \cellcolor{green!10}1.30\\
        3 & Botnet Detection Techniques & 142 & 1.62\\
        \cellcolor{green!10}4 &\cellcolor{green!10} Malware Feature Selection And Extraction &\cellcolor{green!10} 353 & \cellcolor{green!10}4.02\\
        5 & Network Intrusion Detection & 134 & 1.52\\
        \cellcolor{green!10}6 &\cellcolor{green!10} Evaluation of Malware Classifiers &\cellcolor{green!10} 301 & \cellcolor{green!10}3.42 \\
        7 & Malicious Code Analysis & 827 & 9.41\\
        \cellcolor{green!10}8 &\cellcolor{green!10} Artificial Intelligence for Malware & \cellcolor{green!10}888 & \cellcolor{green!10}10.10\\
        9 & Nonnegative Matrix Decomposition & 520 & 5.92\\
        \cellcolor{green!10}10 &\cellcolor{green!10} Security Threat Mitigation& \cellcolor{green!10}180 & \cellcolor{green!10}2.05\\
        11 & Deep Learning for Malware&  113 & 1.29\\
        \cellcolor{green!10}12 &\cellcolor{green!10} Machine Learning Techniques & \cellcolor{green!10}275 & \cellcolor{green!10}3.13\\
        13 & Education Technology& 447 & 5.09 \\
        \cellcolor{green!10}14 &\cellcolor{green!10} Unsupervised Anomaly Detection& \cellcolor{green!10}372 & \cellcolor{green!10}4.23\\
        15 & Ransomware Prevention& 147 & 1.67\\
        \cellcolor{green!10}16 &\cellcolor{green!10} Temporal Graph Forecast& \cellcolor{green!10}307 & \cellcolor{green!10}3.49\\
        17 &  Mobile Malware Detection& 230 & 2.62\\
        \cellcolor{green!10}18 &\cellcolor{green!10}  Adversarial Defense Strategy& \cellcolor{green!10}358 & \cellcolor{green!10}4.07\\
        19 &  IoT Security & 238 & 2.71\\
        \cellcolor{green!10}20 &\cellcolor{green!10} Privacy Protection Challenge& \cellcolor{green!10}628 & \cellcolor{green!10} 7.14\\
        21 &  Sparse Tensor Decomposition& 212 & 2.41 \\
        \cellcolor{green!10}22 &\cellcolor{green!10}  Backdoor Detection& \cellcolor{green!10}350 & \cellcolor{green!10} 3.98\\
        23 &  Neural Network Architecture& 581 & 6.61\\
        \cellcolor{green!10}24 &\cellcolor{green!10} Malware Analysis Techniques &\cellcolor{green!10} 610 & \cellcolor{green!10} 6.94\\
        \hline
  \end{tabular}
\end{table}
\subsection{ Dataset}

Initially, 30 documents specializing on large-scale malware analysis and anomaly detection with tensor decomposition fields were selected by the SME as the core documents to construct the data. These documents were expanded along the citation/reference network 2 times. The final dataset was enumerated at 8,790  scientific publications. From the cleaned corpus, the tensor object was generated.

\subsection{ Extraction of Latent Features}

% \todo[inline]{here please first tell us how many documents were in core dataset, what they were about, and how many it documents it got expanded into.}
After setting up the tensor, the most coherent grouping is determined by iterating through a range of \( k = \{1, 2, 3, \ldots, 45\} \) clusters to decompose. Our analysis determined that 25 topic-clusters represented the optimal division across all evaluated \( k \) values.  The decomposition itself was executed using \textbf{T-ELF} on high-performance computing resources, specifically two AMD EPYC 9454 48-Core Processors. This setup provided a total of 192 logical CPUs, enabling us to complete the entire decomposition process in approximately 2 hours. Following the decomposition, post-processing refined and defined clusters for the topics, which are listed in Table \ref{tab:topic_clusters}.

\subsection{Vector Store}

The 8,790 documents were vectorized and ingested into the Milvus vector store. When questions are posed to the framework, they are also vectorized using this model. Of the total documents, 22\% had full-texts available, which were vectorized into the Milvus. Each document and full-text had a DOI, with the full-texts also including paragraph identifiers.

\begin{figure}[h]
    \centering
    \includegraphics[width=.9\columnwidth]{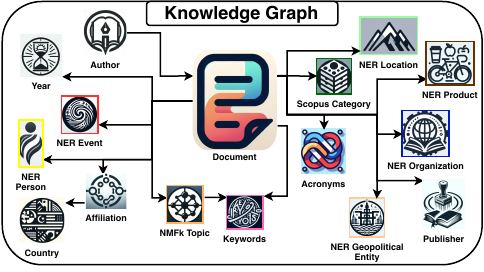}
     \vspace{-1.3em}
    \caption{  The KG schema. 
    % Nodes labels are Document, Author, NERs (Event, Person, Location, Product, Organization, Geopolitical Entity), Publisher, Acronyms, SME Keywords, Affiliation, Affiliation Country, Year, \textbf{T-ELF} Topic, \textbf{T-ELF} Keywords. 
    Images generated with DALL·E \cite{dalle_tensor_decomp_arxiv_images}. }
    \label{fig:KG_SCHEMA}
\end{figure}

\subsection{Knowledge Graph}

\begin{figure}[ht]
    \centering
    \includegraphics[width=\columnwidth]{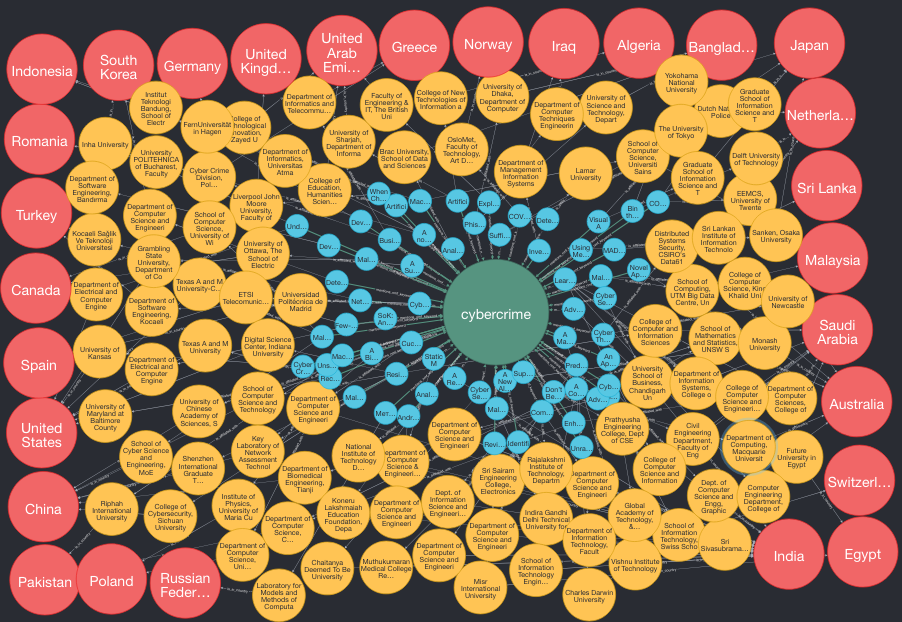}
    \vspace{-2em}
    \caption{  Keyword 'cybercrime' graph search. A single keyword (green), along with linked documents (light blue) are returned. The documents also link affiliated institutions (yellow), and the country of the institutions (red). }
    \label{fig:KG_data}
\end{figure}

From the 25 clusters output form \textbf{T-ELF}, we formatted the the data into 1,457,534 triplets. Once injected into the knowledge graph, there were 321,122 nodes and 1,136,412 edge relationships.  The nodes injected into the graph are represented in Figure \ref{fig:KG_SCHEMA}, where they are organized into 16 base categories, referred to as labels, that define the foundational classes for the injection process. Once the graph was built was directly queried for information as Structured Query Language (SQL) is directly queriable outside of an application. In Figure \ref{fig:KG_data}, the knowledge graph is queried for the SME keyword related to cybercrime. The query is structured as:
 
\begin{center}
\begin{lstlisting}[style=cypherstyle2, xleftmargin=.05\textwidth]
MATCH  (k:Keyword)-[r1]-(d:Document)-[r2]
-(aff:Affiliation)-[r3]-(c:Country )
WHERE k.term CONTAINS 'cybercrime'
RETURN k,r1,d,r2,aff,r3,c
\end{lstlisting}
\end{center}

To retrieve the country nodes from a keyword, several relationships were navigated. First, from the keyword to documents, then from documents to  affiliations and finally from the affiliations to the countries. In the cypher query, these links are the denoted as an r with a following integer, where r is the relationship identifier.  The syntax is ()-[]-()-[]-()-[]-(), where brackets are relations and parenthesis are nodes. In the first part of the ``where" clause, the keyword label is further tailored to the keyword node, such that it must contain ``cybercrime." Overall this can answer the question, ``which countries have published papers that mention cybercrime?"  The question's retrieved nodes in Figure \ref{fig:KG_data} has 29 countries in red, 99 affiliated institutions in yellow, and 65 published documents in blue.

\subsection{Question Answering Validation}

The raw data collected was analyzed using document-specific questions in Zero-Shot Conditioning, including:
\begin{itemize}
\footnotesize
\item How many citations are there for \textit{DOI}?
\item How many references are there for \textit{DOI}?
\item How many authors are there for \textit{DOI}?
\item What year was \textit{DOI} published?
% \item How many affiliated institutions worked on \textit{DOI}?
\item Which publisher published \textit{DOI}?
\item How many scopus categories are assigned to \textit{DOI}?
\item What is the title of \textit{DOI}?
% \item How many countries wrote \textit{DOI}?
% \item Which countries wrote \textit{DOI}?
\end{itemize}
After document specific questions, we then examined topic specific questions, which included year variations, as in: 
\begin{itemize}
\footnotesize
\item How many papers are there on the topic of \textit{Topic}?
\item How many papers were written related to \textit{Topic} in \textit{Year}?
% \item How many papers related to \textit{Topic} are associated with \textit{Country}?
\end{itemize}

\begin{figure}[ht]
    \centering
    \includegraphics[width=\columnwidth]{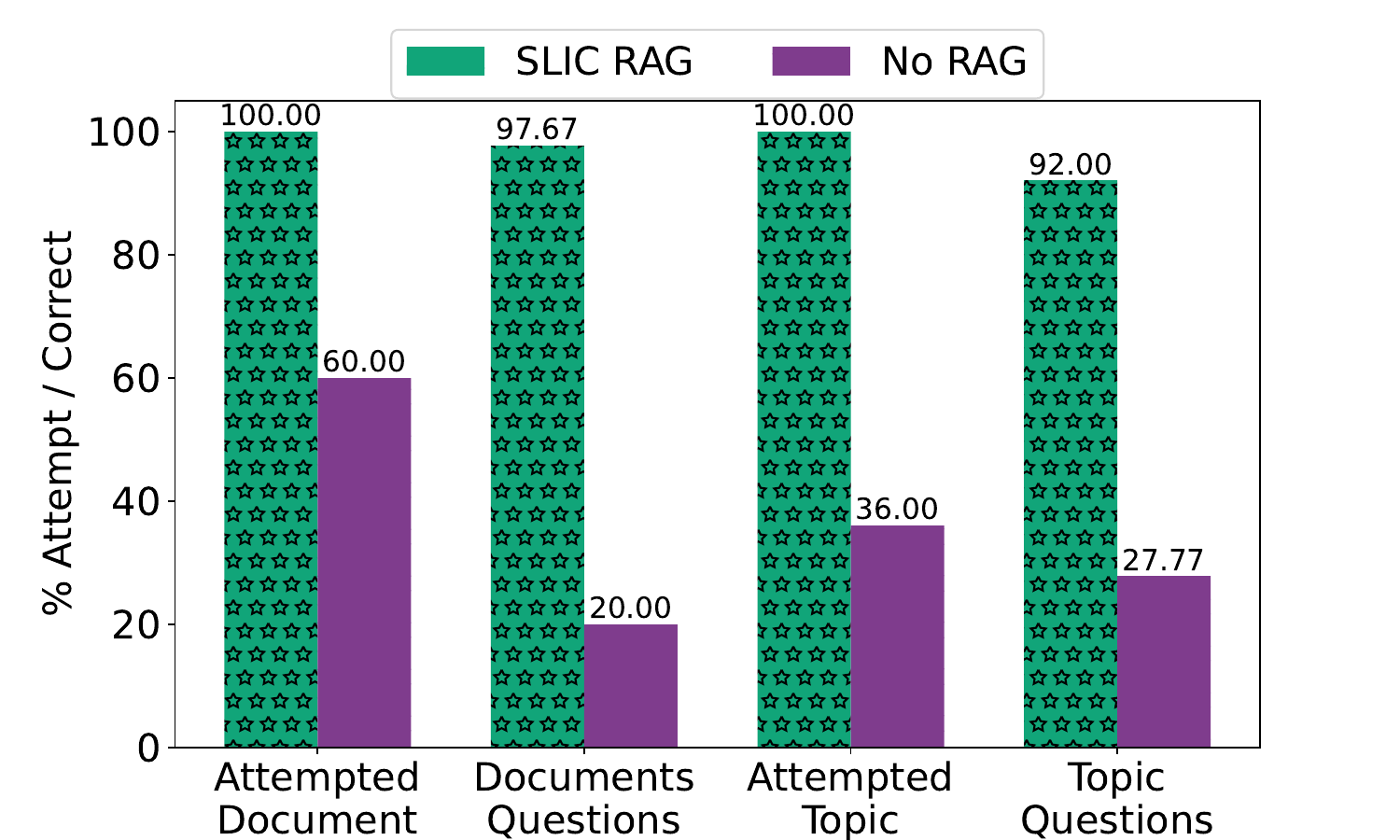}
    \vspace{-2em}
    \caption{ Two question types, document and topic, showing LLM percent attempts and  correct with \textbf{SMART-SLIC} RAG and without RAG. }
    \label{fig:rag_comparison}
\end{figure}

In total, there were 200 questions in this set. Using these questions, in this study, we compare the performance of GPT-4-instruct \cite{openai_api} with and without our RAG framework on both topic-specific and document metadata questions. As shown in Figure \ref{fig:rag_comparison}, our findings indicate that GPT-4 with RAG answers all questions with a 97\% accuracy rate. In contrast, without RAG, GPT-4 abstains from answering 40\% of the questions, and the accuracy of the answered questions drops to 20\%. A similar trend is observed for topic-based questions, where the specialized RAG significantly enhances the retrieval of correct answers. The topic questions attempted with RAG was also 100\%, but without was only 36\%. In consideration of only the attempted questions, the system with RAG answered the topic questions correctly 92\%. Without RAG, the LLM answered the topic questions with 27.77\% accuracy.

Without RAG, several questions about years were answered incorrectly, with the system stating the year didn't exist. The LLM also struggled with author and reference details, often asking for more information or recommending consulting a human expert. In some cases, it noted its lack of internet access but later suggested using Google Scholar, yet still provided inaccurate responses.

The analysis highlights the effectiveness of \textbf{SMART-SLIC} in accurately answering nearly all 200 research queries, covering both document-specific and topic-specific questions. In contrast, GPT-4-instruct without RAG struggled. \textbf{SMART-SLIC} excelled in handling complex queries related to citations, references, authorship, and topical relevance, underscoring the advantages of integrating its methodologies for more reliable AI-driven research tools and improved information retrieval in scientific fields.

We then shifted to more complex questions requiring thorough searches through various retrieval methods, using GPT-4-instruct with and without RAG. The SME posed the following questions, and the responses are from the \textbf{SMART-SLIC} RAG process, with DOI citations provided in square brackets:

% % \begin{enumerate}
%     1) \textit{What challenge is associated with outlier detection in high-dimensional data?}
%         \textcolor{blue}{The challenge associated with outlier detection in high-dimensional data is the ``curse of dimensionality," which makes it difficult to detect outliers in the full data space [10.1109/ICCC54389.2021.9674561].}
        
%     2) \textit{Can tensor decomposition methods be leveraged to detect SCADA anomalies?}
%         \textcolor{blue}{Yes, tensor decomposition methods can be leveraged to detect SCADA anomalies. The use of non-negative tensor decomposition, specifically the Canonical Polyadic Alternating Poisson Regression (CP-APR) method, has been applied to model the complex and multi-faceted activity details of SCADA events, improving the specificity and accuracy of identifying anomalies in electrical grid systems [10.1109/MILCOM58377.2023.10356348].}
        
%     3) \textit{What does MADHAT stand for?}
%          \textcolor{blue}{The acronym MADHAT stands for Multidimensional Anomaly Detection fusing HPC, Analytics, and Tensors [10.1109/HPEC.2019.8916559].}
         
%     4) \textit{What's the most advanced anomaly detection method on the LANL Unified Host and Network Dataset?}
%          \textcolor{blue}{The most advanced anomaly detection method on the Los Alamos National Laboratory Unified Host and Network Dataset is the tensor factorization model that exploits the multi-dimensional and sparse structure of user authentication logs for improved detection of penetration testing events [10.1109/ISI49825.2020.9280524].}
% % \end{enumerate}

    \begin{enumerate}
    \footnotesize  
    \item \textit{What challenge is associated with outlier detection in high-dimensional data?} 
        \textcolor{blue}{The challenge associated with outlier detection in high-dimensional data is the ``curse of dimensionality," which makes it difficult to detect outliers in the full data space [10.1109/ICCC54389.2021.9674561].}
        
    \item \textit{Can tensor decomposition methods be leveraged to detect SCADA anomalies?}
        \textcolor{blue}{Yes, tensor decomposition methods can be leveraged to detect SCADA anomalies. The use of non-negative tensor decomposition, specifically the Canonical Polyadic Alternating Poisson Regression (CP-APR) method, has been applied to model the complex and multi-faceted activity details of SCADA events, improving the specificity and accuracy of identifying anomalies in electrical grid systems [10.1109/MILCOM58377.2023.10356348].}
        
    \item \textit{What does MADHAT stand for?}
        \textcolor{blue}{The acronym MADHAT stands for Multidimensional Anomaly Detection fusing HPC, Analytics, and Tensors [10.1109/HPEC.2019.8916559].}
        
    \item \textit{What's the most advanced anomaly detection method on the LANL Unified Host and Network Dataset?}
        \textcolor{blue}{The most advanced anomaly detection method on the Los Alamos National Laboratory Unified Host and Network Dataset is the tensor factorization model that exploits the multi-dimensional and sparse structure of user authentication logs for improved detection of penetration testing events [10.1109/ISI49825.2020.9280524].}
\end{enumerate}
For each response, the \textbf{SMART-SLIC} agent selected DOIs that the SME also chose, demonstrating the agent's accuracy in retrieving relevant sources. The consistency in DOI selections highlights the robustness of the retrieval mechanisms, ensuring reliable and pertinent information for the user's questions.

The same questions were asked without RAG, and the results varied. The LLM answered the first general question accurately, but while the initial response to the second question was correct, its elaboration missed key details. The third and fourth responses were entirely wrong, with fabricated answers like "Malware and Attack Detection Hunting and Analysis Team" and "Long Short-Term Memory." Additionally, none of the responses included DOI citations, reducing the credibility of the information by omitting source references.

The evaluation of \textbf{SMART-SLIC} and GPT-4-instruct, with and without RAG, highlights the importance of retrieval systems for accurate research output. \textbf{SMART-SLIC}'s RAG excelled in selecting relevant DOI citations for complex queries, while GPT-4-instruct struggled with fabrications, showing the need for advanced systems like \textbf{SMART-SLIC}. Its strength lies in using high-quality, domain-specific corpora for strong performance in defined research areas, while also offering potential for further exploration in less-defined domains.

\section{Conclusion}
\label{sec:conclusion}

Our \textbf{SMART-SLIC} framework leverages advanced language models and specialized tools to effectively address user queries by categorizing them into Specific Document Queries and General Queries for efficient processing. The ReAct agent manages general inquiries, while NER and cypher query generation handle document-specific questions.

LLMs excel in general NLP tasks but struggle in domain-specific areas due to hallucinations, knowledge cut-offs, and lack of attribution. Our system addresses this by integrating RAG with a domain-specific KG and VS, enhancing reliability without fine-tuning. Built using NLP, data mining, and non-negative tensor factorization, this setup enables accurate attributions, reduces hallucinations, and excels in domain-specific queries, as shown in malware analysis research.

The framework significantly enhances query response accuracy and reliability, making it adaptable to various applications. Future work will expand the framework's use across domains like robotics, materials science, legal cases, and quantum computing. Enhancements in graph completion, entity linking, and link prediction will further interconnect graphs, reveal hidden connections, and support LLMs in information clarification, keeping \textbf{SMART-SLIC} at the forefront of intelligent information retrieval and generation.

% \section*{Acknowledgment}
% This research was funded by the U.S. Department of Energy
% National Nuclear Security Administration's Office of Defense Nuclear Nonproliferation Research and Development (DNN R$\&$D).
% Computational resources was provided by the Los Alamos National Laboratory's Institutional Computing Program supported by the U.S. Department of Energy National Nuclear Security Administration under Contract No. 89233218CNA000001.

\bibliographystyle{IEEEtran}
\bibliography{References}

\vspace{12pt}

\end{document}